%% file: jurix2026-machina-pauliana.tex
\documentclass{IOS-Book-Article}

\usepackage{amsmath}
\usepackage{amssymb}
\usepackage{graphicx}
\usepackage{listings}
\usepackage{booktabs}
\usepackage{makecell}
\usepackage{multirow}
\usepackage{float}
\usepackage{subcaption}

\usepackage{mathptmx}
\usepackage{soul}\setuldepth{article}

\def\hb{\hbox to 11.5 cm{}}

\begin{document}

\pagestyle{headings}
\def\thepage{}
\begin{frontmatter}              

\title{Machina Pauliana:\\A neuro-symbolic approach to the actio pauliana in Dutch insolvency law}


\author[A]{\fnms{Daan} \snm{Boven}\orcid{0009-0003-5031-9634},%
  \thanks{Corresponding author: Daan Boven, daan.boven@student.uva.nl}}
\author[A]{\fnms{Gregor} \snm{Behnke}\orcid{0000-0002-1445-9934}}
and
\author[A]{\fnms{David} \snm{Graus}\orcid{0000-0002-6245-0870}}


\address[A]{University of Amsterdam}

\begin{abstract}
Applying a legal provision frequently requires both structured rule-based reasoning and sensitivity to open-textured terms developed in case law. The actio pauliana of Article 42 of the Dutch Bankruptcy Act (Faillissementswet, Fw), which allows an insolvency practitioner to avoid transactions that prejudice creditors, requires both. We present a neuro-symbolic model of Article 42 Fw. Its symbolic core is written in Catala in a literate style that keeps statute and formalisation aligned and was developed through pair programming, with open-textured terms represented as boolean inputs where neural components couple to the symbolic solver. A neural routine structures a case’s natural-language facts into boolean and date inputs. Neural subroutines resolve the open-textured terms through case-based reasoning over Dutch civil judgments, retrieved through a pipeline that filters on the requirement being resolved and expands along the procedural chain of a case. Segmenting and labelling those judgments attains high precision and recall against a manually annotated gold standard of 100 judgments. On ten hand-annotated cases the neuro-symbolic model handles most criteria reasonably but struggles with the knowledge requirement. Retrieval depth shows no clear quantitative effect, although retrieved judgments provide explicit grounding in prior case law.
\end{abstract}

\end{frontmatter}

\input{sections/01-intro}
\input{sections/02-background-legal}
\input{sections/03-background-computational}
\input{sections/04-method}
\input{sections/05-evaluation}
\input{sections/06-conclusion}

\acknowledgements
We thank Henk Jan Boven, insolvency practitioner, and Gert-Jan Boon, assistant professor of company and insolvency law at Leiden University, for their valuable legal expertise and guidance throughout this research.

David Graus is partly funded by ICAI (AI for Open Government Lab).
Views expressed in this paper are not necessarily shared or endorsed by those funding the research.

\bibliographystyle{vancouver}
\bibliography{references}

\end{document}

%% file: sections/01-intro.tex
\section{Introduction}

Research in artificial intelligence and law dates back to the 1970s, and has since often focused on symbolic, rule-based methods, which normalise legal texts into logic-based models that are also intelligible to computers~\cite{Rissland2003AILaw,Ashley2017Chapter2}. 
Today, increasing attention is devoted to data-driven methods, in which AI is used for the automated interpretation of legal texts, document analysis, and the prediction of legal outcomes. 

We apply a hybrid neuro-symbolic model to the avoidance action in Dutch insolvency law (the \textit{actio pauliana}) as codified in Article 42 of the Dutch Bankruptcy Act (Faillissementswet, Fw). 
This provision allows an insolvency practitioner to avoid legal acts (transactions) that cause prejudice to creditors and decrease the estate. 
If creditors are disadvantaged in their possibilities of recovery as a result of legal transactions by the debtor, such transactions may, under certain conditions, be avoided by the insolvency practitioner for the benefit of the estate \cite{DeWeijsFaillissementspauliana2010}. 
Consider, e.g., a director who sells the company car to himself, months before the company is declared bankrupt, by offsetting the car's price against a loan he claims to have granted the company. 
An insolvency practitioner may avoid such a transaction: the sale was voluntary, it allowed the director to recover his own claim in full at the expense of the remaining creditors, and, as it took place within one year of the bankruptcy, knowledge of this prejudice is presumed. 

While the legal structure of Article 42 Fw lends itself to formalisation in terms of logical rules, interpreting several of its open-textured terms requires reference to case law~\cite{Araszkiewicz_2025}.
For that reason we use neural subroutines that resolve open-textured terms through case-based reasoning (CBR) over a corpus of publicly available judgments sourced from rechtspraak.nl (the Rechtspraak corpus).

This paper’s contribution is a full neuro-symbolic workflow for Article 42 Fw: a Catala formalisation~\cite{Merigoux2021Catala} coupled to neural extraction and case-based reasoning over a segmented, labelled, and procedurally linked corpus of Dutch judgments.\footnote{Code, prompts, and data are available at: \url{https://doi.org/10.5281/zenodo.23469695}}

%% file: sections/02-background-legal.tex
\section{Background: Article 42 Fw}
\label{sec:legal-background}

\begin{figure}[]
  \centering
  \includegraphics[width=12cm]{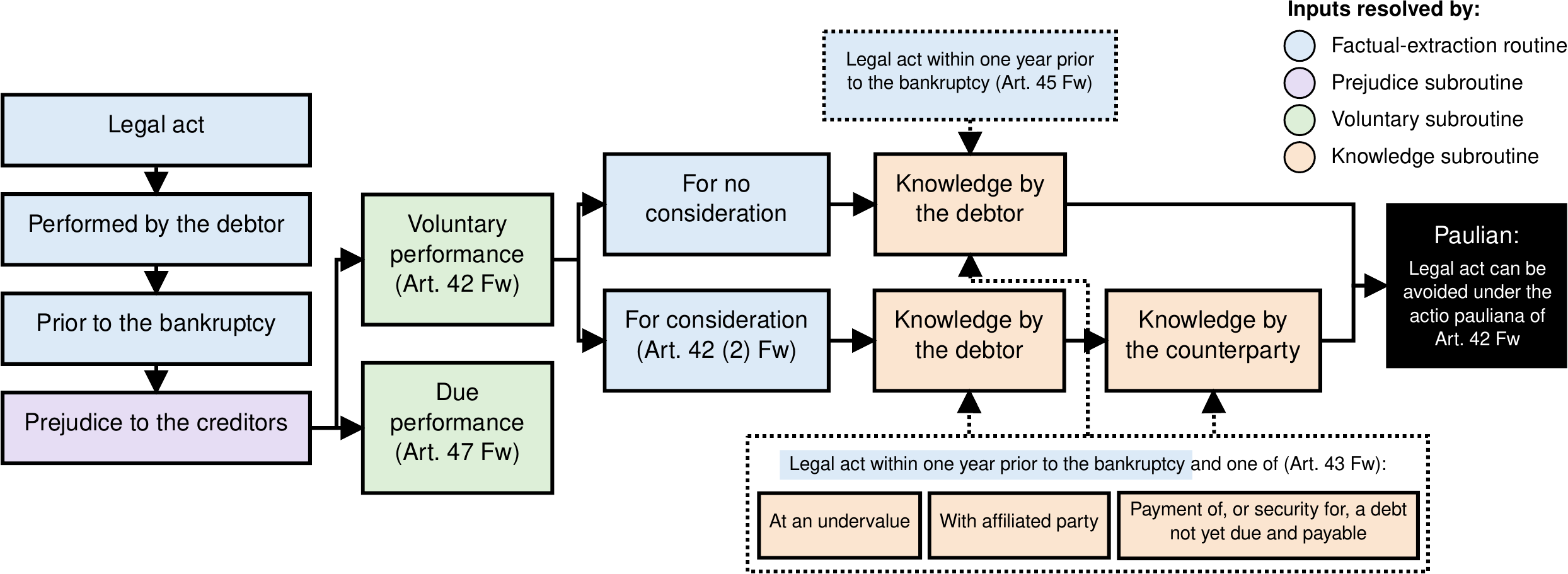}
  \caption{Actio pauliana model based on Pannevis para. 7.2.1 \cite{PannevisHoofdstuk2026}, translation based on Warendorf \cite{WarendorfDutch2015}.}
  \label{fig:actiopaulianapannevisengels}
\end{figure}

\paragraph{Legal definition and description.}
Upon bankruptcy, control of the debtor's estate shifts to the insolvency practitioner, whose powers include avoiding legal acts that prejudice creditors through the \textit{actio pauliana}~\cite{DeWeijsFaillissementspauliana2010,PannevisHoofdstuk2026}.
The insolvency practitioner can invoke the actio pauliana by an extrajudicial declaration, and in practice only contested cases reach the courts~\cite{PannevisHoofdstuk2026,VriesendorpInsolventierecht2024}.
We focus on the avoidance of voluntary legal acts under Article~42~Fw, together with Articles~43 and 45~Fw, which provide evidentiary presumptions that assist the insolvency practitioner. 
Article~42~Fw can be decomposed into five principal elements.
The transaction:
(a) is a legal act, 
(b) was voluntary, 
(c) resulted in prejudice to the general body of creditors,
(d) the debtor knew or ought to have known that the legal act would result in prejudice to the creditors, and 
(e) where the legal act was for consideration, the counterparty also knew or ought to have known that the legal act would result in prejudice to the creditors~\cite{VriesendorpInsolventierecht2024}. 
We visualise these elements as part of the broader decision scheme of the \textit{actio pauliana} in Figure~\ref{fig:actiopaulianapannevisengels}.

\paragraph{Open-textured terms.}
\label{subsec:open-textured-elements}
Some requirements of Article~42~Fw are open-textured and fact-sensitive, and are not suitable for purely symbolic representation. 
For those elements (green, purple, and orange boxes in Figure~\ref{fig:actiopaulianapannevisengels}), case law is not just illustrative but necessary to understand how the rule operates in practice. We address the following terms: 

\textbf{Voluntary performance:}
A legal act is voluntary where it is performed without a legal obligation arising from statute or contract, even where the factual circumstances left the debtor practically no alternative but to perform it. Case law further illustrates the boundary between legal obligation and mere factual or economic necessity \cite{PannevisHoofdstuk2026,WesselsGevolgen2019}.

\textbf{Prejudice to creditors:}
Prejudice is assessed when the actio pauliana is invoked, or, if contested in court, when the court decides on it \cite{WesselsGevolgen2019}. 
At that time, creditors' actual recovery position is compared with the hypothetical situation in which the contested transaction had not occurred. 
This hypothetical does not concern all possible alternatives, but what would most likely have happened had the transaction been omitted \cite{VriesendorpInsolventierecht2024}. 
Case law illustrates how this comparison is applied across different transactions.

\textbf{Knowledge of prejudice:} 
The knowledge requirements of the debtor and counterparty have been objectified by equating actual knowledge with an `ought to have known' standard, so that the insolvency practitioner does not need to prove the debtor's or counterparty's actual state of mind \cite{VriesendorpInsolventierecht2024,WesselsGevolgen2019}. 
The Supreme Court further specified that knowledge is present when, at the time of the transaction, bankruptcy and a deficit in the estate were foreseeable with a reasonable degree of probability \cite{PannevisHoofdstuk2026,HR_2009_Van_Dooren_ABN_AMRO_III}.
Articles~43 and 45~Fw further assist the insolvency practitioner through rebuttable presumptions of knowledge. For acts within one year before bankruptcy, Article~43 presumes knowledge for specified transactions for consideration, while Article~45 does so for acts for no consideration~\cite{PannevisHoofdstuk2026}.

%% file: sections/03-background-computational.tex
\section{Background: computational approaches}

\paragraph{Executable law and Catala.} 
The ``rules-as-code'' paradigm builds on the idea that normalised, machine-executable versions of law can improve transparency, consistency, and traceability between legal text and implementation~\cite{Ashley2017Chapter2}. Tax law is particularly suited to this approach because its rules are comparatively calculative and contain fewer open-textured terms; Dutch initiatives such as \emph{RegelRecht} and \emph{Regelspraak} illustrate this in practice~\cite{RegelRecht2025,Regelspraak2025}, whereas Dutch private and insolvency law remain largely unexplored. Catala is a language for executable law that has been applied primarily to tax law. It brings a systematic approach in which literate programming (annotating the law line by line) and pair programming (a computer scientist and a legal scholar working side by side) are the norm \textit{ex ante}~\cite{Merigoux2021Catala,Huttner2022Catala}. Symbolic approaches, however, may struggle with vagueness and semantic or syntactic ambiguities~\cite{Ashley2017Chapter2,Allen1978NormalizedLegalDrafting}.

\paragraph{Generative AI and Retrieval-Augmented Generation.}
These limitations motivate neural components that can reason over open-textured legal concepts. Large Language Models (LLMs), however, introduce their own limitations, including hallucination, poor calibration, and static knowledge~\cite{JurafskyMartin2026SLP_ch7}. Retrieval-augmented generation (RAG) addresses part of this problem by combining a model's \textit{parametric} memory with the \textit{non-parametric} memory of a retrieved document collection~\cite{Lewis2021RAG}. This early approach, now commonly termed \textit{Naive RAG}, has limitations such as over-reliance on or sensitivity to retrieved text. \textit{Advanced RAG} addresses these by adding pre- and post-retrieval stages, such as metadata filtering, re-ranking, and context compression~\cite{Gao2024RAGSurvey}. Its effectiveness also depends on how documents are segmented: semantic chunking preserves meaningful sections, which aligns well with the structure of judicial decisions. RAG has accordingly been applied to legal question answering~\cite{Wiratunga2024CBRRAG} and legal advice generation~\cite{VanDerMeerRossi2026LegalCheck}, while LLMs have also been used for legal information extraction and document segmentation~\cite{Li2025LEE}.

\paragraph{Neuro-symbolic Legal AI.}
To address LLMs' shortcomings, we draw on Kautz's type 2 Symbolic[Neuro] architecture~\cite{Kautz2022ThirdAISummer}, increasingly applied in legal AI~\cite{Hude2025KnowledgeGraphs,vonCossel2026Rulemapping}:
a symbolic solver is the primary system, and neural components supply structured inputs and resolve open-textured terms through retrieval-augmented case-based reasoning.

The first neuro-symbolic legal system, Split-Up (1995), addresses the Australian Family Law Act 1975 by combining rule-based and neural reasoning across separate subtasks~\cite{Zeleznikow1995SplitUpSystem}, much like our routines. Its networks, however, are trained on past judgments and cannot explain their inferences, so Toulmin argument structures provide explanations separately. Our LLM routines can produce textual explanations grounded in retrieved case law.
More recently, Kant et al.~\cite{Kant2025LogicalLLMsLaw} prompt LLMs to encode insurance contracts and claims in Prolog, finding that structured guidance substantially improves the generated encodings. There the LLM writes the symbolic program, whereas our Catala model is written by hand and the LLM only resolves its inputs.
Closest to our method, Sadowski and Chudziak~\cite{Sadowski2025} propose a three-step process of term identification, predicate extraction, and rule application through symbolic verification, applied to hearsay formalised in first-order logic, which outperforms few-shot prompting. Their rule is likewise defined by experts but their LLM determines predicates from the facts alone, while our CBR subroutines additionally rely on retrieved judgments to resolve open-textured terms.

%% file: sections/04-method.tex
\section{Method}

We propose an approach in which a symbolic problem solver is the primary system. Its structured inputs are derived from the target case’s natural-language facts by neural components: one factual-extraction routine and three RAG-powered CBR subroutines that resolve the open-textured terms identified in Section~\ref{sec:legal-background}.

\subsection{Symbolic model}
\label{sec:symbolic-model}
Our symbolic core is a machine-readable model of Articles~42, 43, and 45~Fw that, given input variables describing a contested transaction, decides whether the transaction is avoidable. 
The model is written in Catala, with the decisive advantage here being its support for pair programming between programmers and lawyers. Our model was developed with the help of an insolvency practitioner and a legal scholar.
Building the model follows the normalisation Allen and Engholm \cite{Allen1978NormalizedLegalDrafting} described for statutes: atomic propositions are identified, replaced by labels, made explicit with propositional logic, and the original text restored to validate the result, in predicate form where possible~\cite{Ashley2017Chapter2}.
Normalisation also exposed several syntactic ambiguities; for instance, we model the evidentiary presumption of Article 43 Fw as rebuttable per side. Semantic ambiguities include counting one year back from 29 February, where we round down to 28 February.

The symbolic model follows the standard decision scheme of the actio pauliana, as outlined in Figure~\ref{fig:actiopaulianapannevisengels}. Some elements, namely whether an act is voluntary, prejudice to creditors, and the knowledge requirement, require more complex case-law-dependent reasoning. Rather than forcing them into further symbolic rules, the model represents them as boolean \textit{inputs} whose truth values are resolved by neural subroutines.

\subsection{Corpus construction}
\label{sec:corpus-construction}
The CBR subroutines rely on an advanced RAG approach, which retrieves relevant segments of judgments from a corpus of judicial decisions concerning the actio pauliana under Article~42~Fw. 
The judgments are sourced from \href{https://rechtspraak.nl}{rechtspraak.nl}, the official open data publication channel of the Dutch judiciary \cite{IvoRechtspraakOpen2019}.

The initial document selection for corpus construction is made in two steps. 
First, we extract a metadata dump from all published judicial decisions through \href{https://woozm.nl}{WooZM}, a publicly accessible Dutch open government search engine. 
We select from this dump all decisions whose \emph{area of law} metadata is classified as civil law.
This yields over 270{,}000 civil-law decisions as XML documents.
Next, we apply regular expressions to filter for Articles~42, 43, and/or~45~Fw, the latter two containing evidentiary presumptions; we deliberately do not use `(actio) pauliana' as a search term, as it also denotes the civil-law actio pauliana of Article 3:45 BW, which is outside our scope.
This yields our final corpus of 846 XML documents representing judicial decisions that involve the \emph{actio pauliana}. 

\subsubsection{Semantic chunking}
\label{sec:corpus-survey}

Rechtspraak XML is structurally inconsistent~\cite{TjongTjinTai2021}, making its markup insufficient for reliable section segmentation. Existing efforts, however, are either no longer accessible~\cite{Trompper2016} or do not identify the substantive judgment sections required here~\cite{Crombach2024RechtspraakSegmentation}.

We therefore build a domain-specific deterministic segmenter. Its lexicon is derived from an LLM-based corpus survey. A separate few-shot prompt is used per court instance. Dutch civil judgments generally follow a recurring structure grounded in Article~230 Rv and reflected in judicial drafting practice~\cite{ahsmann_weg_naar_civiele_vonnis_2020,lovc_hoven_schrijven_2020}. 
Each prompt receives one judgment in XML and asks the model to identify the opening marker of five recurring sections: \textit{course of proceedings}, \textit{facts}, \textit{dispute}, \textit{grounds of decision}, and \textit{decision}. We use \texttt{mistral-small-3.2-24b-instruct-2506} with temperature 0.1, top-$p$ 0.2, and seed 42. The identified opening markers are manually consolidated and iteratively refined into a lexicon of heading words and regex-to-label rules. The corpus survey thus informs a fixed rule set rather than making downstream segmentation itself LLM-dependent.

The resulting segmenter is a cascade of finite-state transducers in the spirit of FASTUS~\cite{Hobbs1997FASTUS}: (i) \emph{recognition} identifies candidate section boundaries using this lexicon; (ii) \emph{refinement} recovers missed boundaries and removes incorrectly detected ones; and (iii) \emph{closing}, the final stage, pairs each section start with the next and assigns the text between them to the corresponding section. The resulting labelled sections are validated against a manually segmented sample and subsequently serve as the semantic chunks used for retrieval.
We evaluate this segmentation approach in Section~\ref{results-prerequisites}.

\subsection{Neural components}
\label{sec:neuro-symbolic-model}
The neural components comprise one factual-extraction routine and three CBR subroutines, all implemented using LLMs. Each routine uses a system prompt containing the relevant statutory provisions and legal doctrine, while the user prompt provides the facts and dispute of the case to be resolved in natural language.
The routines run in parallel and do not communicate. Their outputs are combined by the Catala symbolic solver.

\paragraph{Structured factual-input extraction.}
This routine transforms unstructured facts of a contested transaction in natural language into structured inputs of the Catala model. 
The factual-extraction routine uses an LLM to extract eight structured case inputs from the target case: 
three boolean inputs (\emph{is legal act}, \emph{performed by the debtor}, and \emph{for no consideration}); 
two dates (the \emph{date of the legal act} and \emph{bankruptcy date}); 
and three textual explanations supporting the extracted boolean values. 

\paragraph{Open-textured term resolution.} 
To resolve the open-textured terms discussed in Section~\ref{sec:legal-background} (\emph{voluntary performance}, \emph{prejudice to creditors}, and the \emph{knowledge requirements}), three RAG-powered CBR subroutines retrieve relevant judgments and return boolean-valued inputs to Catala, alongside a textual explanation of the reasoning. 
The \emph{voluntary-performance} and \emph{prejudice} subroutines each resolve to a single boolean input. 
The \emph{knowledge} subroutine resolves to eight boolean inputs corresponding to the knowledge and evidentiary-presumption branches in Figure~\ref{fig:actiopaulianapannevisengels}: directly established debtor and counterparty knowledge; transaction at an undervalue; payment of, or security for, a debt not yet due and payable; affiliated party; pre-existing obligation before the one-year period; and rebuttal of the debtor- and counterparty-knowledge presumptions.

The RAG embeds judgment sections with \texttt{qwen3-embedding-8b}, a multilingual model previously used for legal example retrieval~\cite{Xue_2026_Legal2LogicICL}, with a 32k-token context window accommodating an entire section as a single chunk~\cite{Zhang_2025_qwen3embedding}.
We score each judgment against the natural-language input of the target case, using the judgment's \textit{facts} and \textit{grounds of decision} sections. Where multiple sections of the same type occur, as in judgments addressing several issues, \textit{maxP} aggregation~\cite{Dai_2019} retains the best-matching chunk for each type before the scores are combined using fixed weights, inspired by the CBR-RAG approach of Wiratunga et al.~\cite{Wiratunga2024CBRRAG}.

\paragraph{Pre-retrieval.}
We go beyond \emph{Naive RAG} by adding the following pre- and post-retrieval steps. 
Each judgment in our corpus carries boolean labels indicating whether it rules on each of the three open-textured terms.
For each subroutine, we restrict retrieval to judgments labelled as addressing the corresponding term, reducing the search space and the probability of retrieving irrelevant chunks.

To select a model for labelling our corpus, we manually annotated a gold standard of 100 judgments for the presence of each of the three elements. 
We compared \texttt{gpt-5.4} to \texttt{mistral-medium-3.5-128b:fp8}, and picked the latter as it achieved slightly higher recall (Table~\ref{tab:annotation-results}), which we prefer over precision as missing a relevant case is costlier than retrieving a less directly relevant one, in addition to the open-weights nature of the model, which is an advantage for reproducibility. 

\paragraph{Post-retrieval.}
Higher instance courts frequently rely on facts established in lower courts without restating them.
At the same time, a lower-instance decision may have been overruled, with the higher-instance decision establishing the final reasoning and decision.
In post-retrieval we therefore expand the result set along the procedural chain: when a retrieved decision belongs to a case litigated at another instance, those decisions are added to the retrieved context supplied to the CBR subroutine.

\paragraph{Parameters used in neural components.}
For reproducibility, the inference parameters were held constant where supported.
For \texttt{mistral-medium-3.5-128b:fp8}, used in both the pre-retrieval annotation and the neural routines, we set the temperature to 0.1, top-$p$ to 0.2, and the seed to 42; top-$k$ was not set. 
For \texttt{gpt-5.4}, used in the pre-retrieval annotation experiment, and \texttt{gpt-5.5}, used in the neuro-symbolic model, we set the seed to 42 and reasoning effort to medium. 
For \texttt{gemini-3.5-flash}, used in the neuro-symbolic model, we set the temperature to 0.1, top-$p$ to 0.2, and reasoning effort to medium. 
Parameters not mentioned were left at their default values. 

%% file: sections/05-evaluation.tex
\section{Results}
Each component is evaluated by precision, recall, and $F_1$ against a gold standard annotated manually by the first author.
Due to practical constraints that we discuss in Section~\ref{sec:conclusion}, the end-to-end model is evaluated on a small set of hand-annotated cases. 
For that reason, we also report qualitative findings alongside the quantitative metrics.

\subsection{Prerequisites}
\label{results-prerequisites}
\paragraph{Document segmentation.}
First, we evaluate the document segmenter against a manually annotated random sample of 100 judgments, consisting of 582 segments. 
As Table~\ref{tab:segmentation-results} shows, the segmenter attains high precision, recall, and $F_1$ across all five target sections, with a micro average of $0.98$ precision and $0.97$ recall; the lowest per-section score is on the \textit{grounds of decision} section ($F_1 = 0.94$). 
The segmenter was built through a corpus survey specifically for judgments concerning the Dutch transaction avoidance under Art.~42~Fw, so these results speak only to the \textit{actio pauliana} of Art.~42~Fw, and do not necessarily extend to judgments in other areas of civil law. 

\begin{table}[H]
  \centering
  \setlength{\tabcolsep}{2.5pt}
  \renewcommand{\arraystretch}{0.95}
  \caption{Section segmentation performance on the Rechtspraak corpus, reporting precision, recall, and F1 per section against 100 manually segmented judgments.}
  \label{tab:segmentation-results}
  \begin{tabular}{@{}lrrr r@{\hspace{1em}}lrrr r@{}}
    \toprule
    & \multicolumn{3}{c}{Score}
    & & & \multicolumn{3}{c}{Score} & \\
    \cmidrule(lr){2-4}\cmidrule(lr){7-9}
    Section & P & R & F1 & \ensuremath{n}
      & Section & P & R & F1 & \ensuremath{n} \\
    \midrule
    \textit{Course of proceedings}
      & 1.00 & 0.99 & 0.99 & 100
      & \textit{Established facts}
      & 1.00 & 0.97 & 0.99 & 76 \\
    \textit{Dispute}
      & 0.98 & 0.94 & 0.96 & 63
      & \textit{Grounds of decision}
      & 0.95 & 0.92 & 0.94 & 114 \\
    \textit{Decision}
      & 0.99 & 0.99 & 0.99 & 101
      & \textit{Other}
      & 0.98 & 1.00 & 0.99 & 128 \\
    \midrule
    \multicolumn{1}{@{}l}{Micro average (all)}
      & 0.98 & 0.97 & 0.98 & 582
      & \multicolumn{5}{l@{}}{} \\
    \bottomrule
  \end{tabular}
\end{table}

\paragraph{Document labelling.}
Accurate segmentation of judicial decision documents enables the grounds of the decision and the core reasoning by the court to be annotated for whether the court ruled on one of the three open-textured legal criteria presented in Section~\ref{sec:legal-background}. 
We apply an LLM to label document segments with whether the court rules on these (voluntary performance, prejudice, and the knowledge requirements) in the grounds of its decision. 
We evaluate them against the same sample of 100 documents used for evaluating the segmenter, which we also manually annotated for these labels; the results are presented in Table~\ref{tab:annotation-results}. 
We report precision, recall, and $F_1$-score, but an important focus here is on recall, since for jurists the cost of missing information in a legal context is substantially higher than the cost of reviewing extra, irrelevant results. 

The two models seem to trade precision against recall: Mistral reaches higher recall (micro-average $0.94$) at lower precision ($0.80$), whereas GPT-5.4 achieves higher precision ($0.93$) at the cost of slightly lower recall ($0.88$), with micro-averaged $F_1$ scores of $0.86$ and $0.91$ respectively. 
Given the importance of recall and the greater reproducibility and portability of an open-weight model, we select Mistral, which misses the fewest relevant rulings, whereas GPT-5.4 returns fewer false positives.

\begin{table}[H]
  \centering
  \setlength{\tabcolsep}{2.5pt}
  \renewcommand{\arraystretch}{0.95}
  \caption{Pre-retrieval annotation performance per element against 100 manually annotated judgments.}
  \label{tab:annotation-results}
  \begin{tabular*}{\linewidth}{@{\extracolsep{\fill}}lrrrr@{\hspace{1em}}rrrr@{}}
    \toprule
    & \multicolumn{4}{c}{\makecell{\texttt{mistral-medium-3.5-128b:fp8}}}
    & \multicolumn{4}{c}{\texttt{gpt-5.4}} \\
    \cmidrule(lr){2-5}\cmidrule(l){6-9}
    Element & P & R & F1 & \ensuremath{n}
      & P & R & F1 & \ensuremath{n} \\
    \midrule
    Voluntary performance
      & 0.69 & 0.97 & 0.81 & 39
      & 0.85 & 0.90 & 0.88 & 39 \\
    Prejudice to creditors
      & 0.89 & 0.91 & 0.90 & 34
      & 1.00 & 0.82 & 0.90 & 34 \\
    Knowledge
      & 0.86 & 0.92 & 0.89 & 39
      & 0.97 & 0.92 & 0.95 & 39 \\
    \midrule
    Micro average
      & 0.80 & 0.94 & 0.86 & 112
      & 0.93 & 0.88 & 0.91 & 112 \\
    \bottomrule
  \end{tabular*}
\end{table}

\subsection{Neuro-symbolic model}

Finally, we evaluate the neuro-symbolic model against ten hand-annotated cases from outside our Rechtspraak corpus and not used in developing the model, drawn mainly from \emph{Jurisprudentie Onderneming \& Recht} (JOR), a Dutch jurisprudence journal that publishes legally noteworthy cases. 
Their facts and the dispute at hand served as the natural-language input for the neuro-symbolic model. 
Our symbolic Catala model computes a single output from a given set of inputs (truth values of legal criteria, and dates).
To provide a more meaningful comparison, we compare both the model's output and its inputs against the gold standard.
We additionally vary the number $k$ of documents retrieved for case-based reasoning.
We report micro-averaged precision, recall, and $F_1$ for different values of $k$.
Because the validation set is small and consists largely of cases selected for publication in a jurisprudence journal, the results should be interpreted as indicative rather than conclusive.
In Table~\ref{tab:neurosymbolic-results}, we compare three LLMs for the end-to-end neuro-symbolic approach.
Micro-averaged scores for all three models fall in a comparable range (precision between $0.63$ and $0.68$, recall between $0.82$ and $0.90$, and $F_1$ between $0.72$ and $0.77$).
The retrieval depth does not meaningfully affect these aggregate figures: for each model the scores at $k=0$, $k=5$, and $k=10$ are similar, even though the retrieved decisions do visibly change the reasoning the model puts forward for its answers.
Neither a clear ranking between the models nor a clear effect of retrieval depth can be drawn from these figures.

\begin{table}[H]
  \centering
  \setlength{\tabcolsep}{5pt}
  \caption{Neuro-symbolic model performance, reporting micro-averaged precision, recall, and $F_1$ of the boolean criteria and the final verdict over ten judgments against the gold standard, at retrieval depths $k=0$, $k=5$, and $k=10$.}
  \label{tab:neurosymbolic-results}
  \begin{tabular}{lrrrrrrrrr}
    \toprule
    & \multicolumn{3}{c}{$k = 0$} & \multicolumn{3}{c}{$k = 5$} & \multicolumn{3}{c}{$k = 10$} \\
    \cmidrule(lr){2-4}\cmidrule(lr){5-7}\cmidrule(lr){8-10}
    Model & P & R & $F_1$ & P & R & $F_1$ & P & R & $F_1$ \\
    \midrule
    \texttt{gemini-3.5-flash}            & 0.64 & 0.88 & 0.74 & 0.67 & 0.90 & 0.77 & 0.63 & 0.88 & 0.74 \\
    \texttt{gpt-5.5}                     & 0.68 & 0.90 & 0.77 & 0.67 & 0.90 & 0.77 & 0.67 & 0.90 & 0.77 \\
    {\texttt{mistral-medium-3.5-128b:fp8}} & 0.66 & 0.82 & 0.73 & 0.65 & 0.82 & 0.72 & 0.66 & 0.82 & 0.73 \\
    \bottomrule
  \end{tabular}
\end{table}

Retrieval can provide explicit textual grounding in prior judgments even where aggregate scores remain similar. In the validation case \href{https://deeplink.rechtspraak.nl/uitspraak?id=ECLI:NL:RBOVE:2026:4561}{ECLI:NL:RBOVE:2026:4561}, GPT-5.5 at $k=5$ supports its finding that the payment was voluntary by citing \href{https://deeplink.rechtspraak.nl/uitspraak?id=ECLI:NL:GHSHE:2006:BC2959}{ECLI:NL:GHSHE:2006:BC2959}, ``where performance of a natural obligation does not remove the voluntary character of the legal act.'' 
This reference supports treating a gift as a voluntary legal act even if it fulfilled a natural obligation, as the court explains in paragraph 7.5 of the retrieved judgment. 
Since GPT-5.5 already reaches the same conclusion at $k=0$, the visible addition is primarily explicit, verifiable grounding in prior case law. 

In another case, JOR 2026/137, GPT-5.5 at $k=5$ cites \href{https://deeplink.rechtspraak.nl/uitspraak?id=ECLI:NL:GHLEE:2007:BB3627}{ECLI:NL:GHLEE:2007: BB3627} to support its finding that the onward payment to a third party fulfilled an existing obligation. 
The court itself refers to the same judgment as a ``somewhat comparable case'' in footnote 1 to paragraph 3.4. 
The retrieval step of the neuro-symbolic model thus surfaces a legally relevant case that the court itself also cites in the gold standard.

To better understand the model's performance, we break down the results per CBR subroutine in Table~\ref{tab:subroutine-results}. 
The voluntary-performance and prejudice subroutines reach high recall and moderate-to-high $F_1$ across all three models, whereas the knowledge subroutine attains markedly lower precision, recall, and $F_1$ throughout.
For the knowledge subroutine, only Gemini scores slightly higher at $k=5$ than at $k=0$. The other two models remain largely unchanged, so the retrieved prior decisions do not systematically raise performance on this element.

\begin{table}
  \centering
  \setlength{\tabcolsep}{5pt}
  \caption{Per-subroutine micro-averaged precision, recall, and $F_1$ at retrieval depths $k=0$, $k=5$, and $k=10$.}
  \label{tab:subroutine-results}
  \begin{tabular}{llrrrrrrrrr}
    \toprule
    & & \multicolumn{3}{c}{$k = 0$} & \multicolumn{3}{c}{$k = 5$} & \multicolumn{3}{c}{$k = 10$} \\
    \cmidrule(lr){3-5}\cmidrule(lr){6-8}\cmidrule(lr){9-11}
    Model & Subroutine & P & R & $F_1$ & P & R & $F_1$ & P & R & $F_1$ \\
    \midrule
    \multirow{3}{*}{\texttt{gemini-3.5-flash}}
      & Voluntary   & 0.78 & 0.88 & 0.82 & 0.88 & 0.88 & 0.88 & 0.78 & 0.88 & 0.82 \\
      & Prejudice  & 0.67 & 1.00 & 0.80 & 0.60 & 1.00 & 0.75 & 0.67 & 1.00 & 0.80 \\
      & Knowledge               & 0.10 & 0.40 & 0.16 & 0.16 & 0.60 & 0.25 & 0.10 & 0.40 & 0.16 \\
    \cmidrule(l){1-11}
    \multirow{3}{*}{\texttt{gpt-5.5}}
      & Voluntary   & 0.88 & 0.88 & 0.88 & 0.88 & 0.88 & 0.88 & 0.88 & 0.88 & 0.88 \\
      & Prejudice  & 0.67 & 1.00 & 0.80 & 0.75 & 1.00 & 0.86 & 0.75 & 1.00 & 0.86 \\
      & Knowledge               & 0.16 & 0.60 & 0.25 & 0.14 & 0.60 & 0.23 & 0.14 & 0.60 & 0.23 \\
    \cmidrule(l){1-11}
    \multirow{3}{*}{ \makecell[l]{\texttt{mistral-medium-} \\ \texttt{3.5-128b:fp8}} }
      & Voluntary   & 0.80 & 1.00 & 0.89 & 0.80 & 1.00 & 0.89 & 0.80 & 1.00 & 0.89 \\
      & Prejudice  & 0.60 & 1.00 & 0.75 & 0.60 & 1.00 & 0.75 & 0.60 & 1.00 & 0.75 \\
      & Knowledge               & 0.07 & 0.20 & 0.11 & 0.07 & 0.20 & 0.10 & 0.07 & 0.20 & 0.11 \\
    \bottomrule
  \end{tabular}
\end{table}

\paragraph{Qualitative inspection} shows differences in how the models reach their conclusions. In JOR 2026/137, GPT-5.5 and Gemini at $k=5$ cite \href{https://deeplink.rechtspraak.nl/uitspraak?id=ECLI:NL:GHSHE:2014:4568}{ECLI:NL:GHSHE:2014:4568} to reach opposite conclusions on prejudice to creditors. 
Gemini follows the insolvency practitioner’s argument: the onward payment to a third party removes funds available to the general body of creditors, and it applies \href{https://deeplink.rechtspraak.nl/uitspraak?id=ECLI:NL:GHSHE:2014:4568}{ECLI:NL:GHSHE:2014:4568} by analogy. 
GPT-5.5 instead distinguishes the present case from that precedent, treating the onward payment as putting right a mistake made by the bank, and finds no prejudice, in line with the court in the gold standard. 
The same case-law reference can thus, particularly in hard cases, support different but defensible positions grounded in prior judgments.

In JOR 2014/78, all three models at $k=5$ find the debtor's knowledge of prejudice unproven, leading the symbolic model to an incorrect verdict that the transaction is not avoidable. The court instead treats knowledge as established because it was insufficiently disputed, a finding that is hard to infer from the model input. This is consistent with Nouwens et al.'s~\cite{Nouwens_2024} observation that imperfect accuracy may reflect differences between the knowledge available to the system and the human decision-maker. Mistral's explanation adds a second error: it invokes the evidentiary presumption by incorrectly placing the 2005 legal act within one year of the 2009 bankruptcy order. The symbolic model correctly rejects the presumption because the dates are over one year apart, but the explanation is not updated accordingly. Such drift in textual explanations follows from the loose symbolic-neural coupling and could be addressed in future work through chain-of-thought prompting.
The same incorrect verdict thus masks different reasoning failures that boolean scores alone cannot distinguish.

Finally, the accuracy of the final verdict, whether the transaction is avoidable under Art.~42~Fw, is reported in 
Table~\ref{tab:final-verdict-accuracy}
as the number of correct verdicts out of the ten cases. 
The accuracy is modest for this sample. 
Manual inspection of the neuro-symbolic model's output reveals no single recurring cause for a diverging final verdict. For none of the models does $k = 10$ yield more correct verdicts than $k = 5$.

\begin{table}[H]
  \centering
  \caption{Correct final verdicts (avoidable under Article~42~Fw) out of ten gold-standard cases, by model and retrieval depth.}
  \label{tab:final-verdict-accuracy}
  \begin{tabular}{lrrr}
    \toprule
    Model & $k = 0$ & $k = 5$ & $k = 10$ \\
    \midrule
    \texttt{gemini-3.5-flash}             & 7 & 7 & 6 \\
    \texttt{gpt-5.5}                      & 7 & 7 & 7 \\
    \texttt{mistral-medium-3.5-128b:fp8} & 6 & 6 & 6 \\
    \bottomrule
  \end{tabular}
\end{table}

%% file: sections/06-conclusion.tex
\section{Conclusion}
\label{sec:conclusion}
We show how Article 42 Fw can be modelled neuro-symbolically by combining a symbolic Catala model that follows the decision scheme of the actio pauliana with neural routines for factual extraction and open-textured legal reasoning. A pipeline for collecting, segmenting, and LLM-labelling judgments provides the corpus used for retrieval, with both segmentation and labelling performing strongly against our gold standard.

On ten hard, out-of-corpus validation cases, performance of the neuro-symbolic model is strongest for voluntary performance and prejudice to creditors, while the results indicate particular difficulty in resolving the knowledge requirement. Increasing retrieval depth does not yield a clear quantitative improvement, but upon inspection, retrieved judgments do seem to strengthen the model’s textual reasoning through explicit and legally relevant grounding in prior case law.

The present court-centric evaluation captures only one way in which the model may be assessed. The validation set consists largely of cases published in a jurisprudence journal precisely because they are legally interesting, and is therefore skewed towards harder cases.
Uncontested, routine cases are settled by an extrajudicial declaration and therefore leave no public record, so they could not be evaluated. In the harder cases that do reach the courts, a model conclusion that differs from the court may still reflect a defensible argument for the opposing side, as the qualitative inspection of the model’s reasoning illustrates. Future work should therefore evaluate the model with insolvency practitioners, who can steer its inputs and argumentation and thereby enable further validation of its legal reasoning.